\PassOptionsToPackage{table}{xcolor}
\documentclass[sigconf,natbib=true]{acmart}
\newcommand{\blue}[1]{{\color{blue} #1}}

\newcommand{\para}[1]{\vspace{1.5ex}\noindent\textbf{#1}}

\usepackage{xspace}
\usepackage{threeparttable}
\usepackage{multirow}
\usepackage[utf8]{inputenc}
\usepackage[most]{tcolorbox}
\usepackage{booktabs} % For \toprule, \midrule, \bottomrule
\usepackage{caption}  % For better table captions
\usepackage{graphicx} % If needed for images
\usepackage{enumitem}
\usepackage{soul}        % For highlighting
\sethlcolor{yellow}      % Highlight color (change as desired)

\usepackage{pifont}% http://ctan.org/pkg/pifont
\newcommand{\cmark}{\ding{51}}%
\newcommand{\xmark}{\ding{55}}%

\usepackage{placeins} % \FloatBarrier

\newcommand{\method}{\textsf{LaViC}\xspace}

\newcommand{\methodfullul}{\underline{La}rge \underline{Vi}sion-Language \underline{C}onversational Recommendation Framework\xspace}

\AtBeginDocument{%
  }

\begin{document}
%\settopmatter{printacmref=true}

\title{\method: Adapting Large Vision-Language Models to Visually-Aware Conversational Recommendation}

\author{Hyunsik Jeon}
\affiliation{
	\institution{UC San Diego}
	\city{San Diego, CA}
	\country{USA}
}
\email{hyjeon@ucsd.edu}

\author{Satoshi Koide}
\affiliation{
	\institution{Toyota Motor North America, Inc.}
	\city{Ann Arbor, MI}
	\country{USA}
}
\email{satoshi.koide@toyota.com}

\author{Yu Wang}
\affiliation{
	\institution{UC San Diego}
	\city{San Diego, CA}
	\country{USA}
}
\email{yuw164@ucsd.edu}

\author{Zhankui He}
\affiliation{
	\institution{UC San Diego}
	\city{San Diego, CA}
	\country{USA}
}
\email{zhh004@ucsd.edu}

\author{Julian McAuley}
\affiliation{
	\institution{UC San Diego}
	\city{San Diego, CA}
	\country{USA}
}
\email{jmcauley@ucsd.edu}

\renewcommand{\shortauthors}{Jeon et al.}

\begin{abstract}
% Conversational recommendation systems aim to elicit user preferences through interactive dialogues.
Conversational recommender systems engage users in dialogues to refine their needs and provide more personalized suggestions.
Although textual information suffices for many domains, visually driven categories such as fashion or home decor potentially require detailed visual information related to color, style, or design.
To address this challenge, we propose \method (\methodfullul), a novel approach that integrates compact image representations into dialogue-based recommendation systems.
\method leverages a large vision-language model in a two-stage process:
(1) visual knowledge self-distillation, which condenses product images from hundreds of tokens into a small set of visual tokens in a self-distillation manner, significantly reducing computational overhead, and
(2) recommendation prompt tuning, which enables the model to incorporate both dialogue context and distilled visual tokens, providing a unified mechanism for capturing textual and visual features.
To support rigorous evaluation of visually-aware conversational recommendation, we construct a new dataset by aligning Reddit conversations with Amazon product listings across multiple visually oriented categories (e.g., fashion, beauty, and home).
This dataset covers realistic user queries and product appearances in domains where visual details are crucial.
Extensive experiments demonstrate that \method significantly outperforms text-only conversational recommendation methods and open-source vision-language baselines.
Moreover, \method achieves competitive or superior accuracy compared to prominent proprietary baselines (e.g., GPT-3.5-turbo, GPT-4o-mini, and GPT-4o), demonstrating the necessity of explicitly using visual data for capturing product attributes and showing the effectiveness of our vision-language integration.
Our code and dataset are available at \url{https://github.com/jeon185/LaViC}.
\end{abstract}

% \begin{CCSXML}
% <ccs2012>
%    <concept>
%        <concept_id>10002951.10003317.10003347.10003350</concept_id>
%        <concept_desc>Information systems~Recommender systems</concept_desc>
%        <concept_significance>500</concept_significance>
%        </concept>
%  </ccs2012>
% \end{CCSXML}

% \ccsdesc[500]{Information systems~Recommender systems}

% \keywords{Conversational Recommendation, Large Language Models, Large Vision-Language Models}

\maketitle

\section{Introduction}
\label{sec:intro}

% Problem
Conversational recommendation has emerged as a promising framework for e-commerce and digital entertainment, offering an interactive channel for users to express their preferences through natural language~\cite{Christakopoulou16,LiKSMCP18,HeXJSLFMKM23,XieWJHSJLKM24,he25reindex}.
Unlike traditional recommendation approaches that rely primarily on past interactions, conversational recommender systems aim to find relevant and personalized items by engaging in a dialogue with users.
% This interactive approach helps clarify ambiguous requests, capture specific constraints, and ultimately enhance user satisfaction.

% LLMs in CRS
Recently, \emph{large language models} (LLMs)~\cite{openai2022chatgpt,abs-2302-13971,abs-2307-09288,vicuna2023,BrownMRSKDNSSAA20,HeXJSLFMKM23,abs-2308-06212,abs-2305-06474} have demonstrated a strong ability to interpret user intentions within natural language conversational contexts.
Their broad pre-training on extensive tasks allows them to handle complex linguistic features and incorporate specialized domain information when generating responses.
This capability has led LLM-based approaches to outperform previous non-LLM methods in conversational recommendation~\cite{HeXJSLFMKM23, WangTZWW23, abs-2308-06212}, positioning LLMs as an essential component in the design of current conversational recommender systems.

\begin{figure*}[t]
	\centering
	\includegraphics[width=1.0\linewidth]{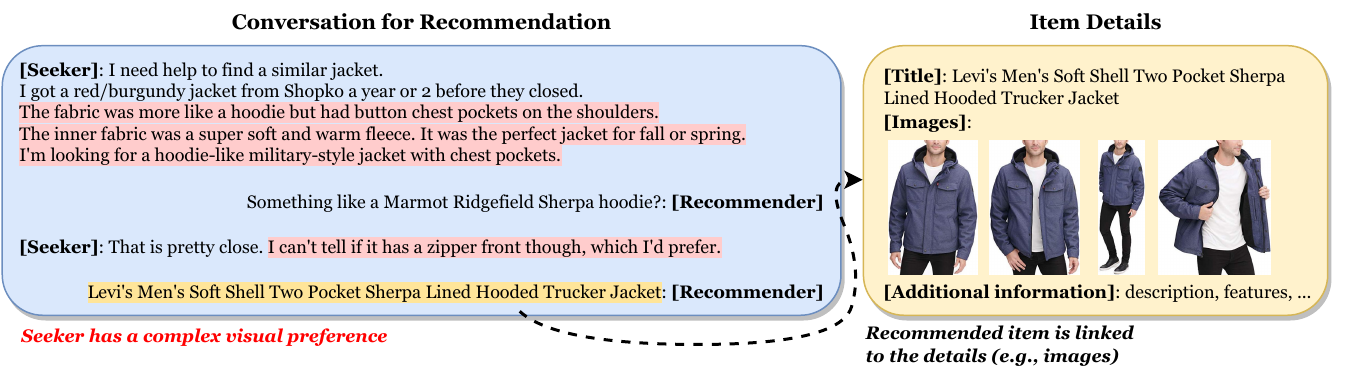}
	\caption{
        The \emph{Reddit-Amazon} dataset contains conversations between a seeker and a recommender. The seeker requests an item recommendation via text, focusing primarily on visual preferences. The item recommended to the seeker by the recommender is linked to detailed item information such as its title and images.
	}
\label{fig:data_illustration}
\end{figure*}

% Motivation
Although text-based recommendation often suffices for many domains, certain product categories (e.g., fashion and home decor) rely heavily on visual features such as color, style, and overall design.
% Purely textual descriptions may not capture these subtle aesthetic or functional aspects.
Previous visually-aware recommender systems~\cite{McAuleyTSH15,HeM16a,KangFWM17} have shown that images can substantially improve performance in visually oriented domains, indicating that purely textual descriptions may miss subtle aesthetic or functional details.
For example, a user who requests a \emph{``hoodie-like military-style jacket with chest pockets''} can narrow the options using textual information, yet verifying the exact silhouette or pocket arrangement depends on visual inspection (Figure~\ref{fig:data_illustration}).
Thus, incorporating visual information helps resolve ambiguities and improves recommendation quality, making it more likely that a system identifies items aligned with the intended look or function of the user.
% However, existing LLM-based approaches focus mainly on textual input, leaving visual information underutilized for domains where appearance is pivotal.

% Challenges
To fully integrate visual information into the conversation, one could resort to \emph{large vision-language models} (VLMs)~\cite{LiuLWL23a,liu2023improvedllava,liu2024llavanext} that unify visual and textual inputs.
However, such models generally tokenize each image into hundreds or thousands of tokens, incurring a steep computational load if multiple items are analyzed simultaneously.
Moreover, naive end-to-end fine-tuning of massive multimodal architectures can lead to overfitting, particularly in specialized domains with limited labeled data.
Finally, directly passing all retrieved items as raw image tokens can be prohibitively expensive in real-world deployments, indicating the need for a compact representation of visual content.

% Our work
In this work, we address these challenges with a two-stage framework for \emph{visually-aware conversational recommendation} built upon VLMs.
First, we perform knowledge distillation to compress the high-dimensional encoding of each product image into a minimal set of visual tokens, preserving essential details of appearance while reducing computational overhead.
Second, we adopt a recommendation prompt tuning procedure on a distilled vision-language model to jointly process textual queries and these compact visual tokens within a unified generative paradigm.
This design balances computational feasibility with the need for detailed visual understanding.
Furthermore, to enable rigorous empirical evaluation, we construct a new dataset, called \emph{Reddit-Amazon} dataset, by aligning Reddit dialogues with Amazon product images, reflecting genuine user queries and realistic visual attributes seldom captured in existing benchmarks.

% Contributions
The main contributions of our work are as follows.
\begin{itemize}
    \item We mitigate the token-explosion problem by condensing each product image into a minimal set of visual tokens, preventing the model from processing thousands of image tokens, and thus improving training stability, reducing overfitting, and ultimately enhancing recommendation accuracy.
    \item We propose a prompt-tuning procedure on a distilled vision-language model, enabling it to jointly process textual dialogues and these compressed visual tokens within a unified generative framework for recommendation.
    \item We release \emph{Reddit-Amazon}, a new dataset consisting of over 19K Reddit conversations (51K turns in total) aligned with Amazon product items. It spans three visually oriented categories (beauty, fashion, and home), linking each recommended item to its title and images. This dataset offers a richer testbed for visually-aware conversational recommendation.
\end{itemize}

\section{Preliminaries}
\label{sec:preliminary}

\subsection{Problem Statement}
\label{subsec:problem}
In the \emph{visually-aware conversational recommendation} task, we aim to recommend relevant items to a user (the seeker) through a multi-turn dialogue while considering both textual and visual features of the items.
Let $\mathcal{I}$ be a set of items, where each item $i \in \mathcal{I}$ has a textual title $\mathit{title}_i$ and a single image $\mathit{image}_i$.
A vocabulary $\mathcal{V}$ defines the tokens used for text.
The conversation is $\mathcal{C} = \{ s_t \}_{t=1}^T$, where each utterance $s_t$ is produced by the seeker or the recommender, drawn from $\mathcal{V}$.
The seeker initially requests recommendations, and at each recommender turn $k$, the system produces a ranked list $\hat{\mathcal{I}}_k \subseteq \mathcal{I}$ that aligns with the true set $\mathcal{I}_k$.
This set $\mathcal{I}_k$ embodies the user’s current preferences, which may be represented through the conversation.
In our setting, we assume $|\hat{\mathcal{I}}_k|=|\mathcal{I}_k|=1$, which means there exists a single ground-truth item and we also aim to recommend a single item.

Generative models such as large language models (LLMs) or large vision–language models (VLMs) often produce unconstrained outputs, potentially recommending items that do not exist in $\mathcal{I}$.
To avoid recommending non-existent items, we use a \emph{candidate-based} approach: a retrieval module supplies a small set of candidate items, and the model selects the correct one.
Furthermore, we prioritize \emph{recommendation accuracy} over the generation of fully fluent dialogues, focusing on correct item selection rather than natural-sounding responses as in previous works~\cite{HeXJSLFMKM23,he25reindex}.

\subsection{Large Vision-Language Models}
\label{subsec:lvms}

We provide an overview of large vision-language models (VLMs) (e.g., LLaVA~\cite{LiuLWL23a,liu2023improvedllava,liu2024llavanext}), which serve as the backbone for our visually-aware conversational recommendation.
These models combine a vision encoder (e.g., CLIP~\cite{RadfordKHRGASAM21} or SigLIP~\cite{ZhaiM0B23}) with a large language model (LLM) (e.g., Vicuna~\cite{vicuna2023} or Mistral~\cite{abs-2310-06825}), enabling both image and text inputs to be handled in a unified transformer-based pipeline.

\para{Vision encoder.}
A common approach for the vision encoder relies on Vision Transformers (ViT)~\cite{DosovitskiyB0WZ21}, which divides an input image into $R$ patches.
Then, each patch is flattened and linearly embedded in a $d$ dimensional vector. A special token (e.g., \texttt{[CLS]}) could be appended, giving $R+1$ tokens.
These tokens are fed together into a transformer, yielding contextualized embeddings:
\begin{equation}
    (\mathbf{p}_0,\;\dots,\;\mathbf{p}_R)
    \;\mapsto\;
    (\mathbf{e}_0,\;\dots,\;\mathbf{e}_R),    
\end{equation}
where $\mathbf{p}_0$ and $\mathbf{e}_0$ correspond to the \texttt{[CLS]} token and its embedding, respectively.
Some frameworks (e.g., LLaVA-v1.5~\cite{liu2023improvedllava} and LLaVA-v1.6~\cite{liu2024llavanext}) further subdivide the image into multiple sub-images, but the patch-oriented mechanism remains the same.

\para{Aligning vision encoder with LLM.}
Once the vision encoder produces $\{\mathbf{e}_r\}_{r=0}^R$, a projector $\Omega_{\mathrm{proj}}$ maps these embeddings into the LLM’s space:
\begin{equation}
    \mathbf{v}_r \;=\;
    \mathrm{Proj}(\mathbf{e}_r;\,\Omega_{\mathrm{proj}}),
\end{equation}
where \(\mathbf{v}_r \in \mathbb{R}^d\).
Let $\Omega_{\mathrm{vision}}$ encompass the vision encoder and projector.
The text input $\mathcal{T}$ is tokenized into $\{\mathbf{x}_t\}_{t=1}^{|\mathcal{T}|}$.
Then, LLM processes the following textual and visual embeddings:
\begin{equation}
    [\mathbf{x}_1,\dots,\mathbf{x}_{|\mathcal{T}|},\,
    \mathbf{v}_0,\dots,\mathbf{v}_R],
\end{equation}
with $\Omega_{\mathrm{LM}}$ denoting the LLM parameters.
Over multiple layers of self-attention and feed-forward blocks~\cite{VaswaniSPUJGKP17}, these embeddings merge into a contextualized representation.

\para{Pretraining and fine-tuning.}
The pretraining of VLMs involves image-conditioned text generation~\cite{LiuLWL23a}.
At inference time, VLMs produce a token-level distribution:
\begin{equation}
    P_{\Omega_{\mathrm{LM}} + \Omega_{\mathrm{vision}}}
    (\mathbf{y}\,\mid\,\mathcal{T},\text{Image}),
\end{equation}
where $\mathbf{y}$ are output tokens (e.g., a caption or an answer).
Fine-tuning for downstream tasks (e.g., visual question answering or chatbot) updates either all parameters or a subset thereof via LoRA~\cite{HuSWALWWC22}, preserving the model's pre-trained multimodal knowledge.

\para{Challenges in using multiple images.}
Whereas standard VLM tasks (e.g., captioning) generally involve a single image, visually-aware conversational recommendation may require analyzing multiple items (e.g., ten or more) in one query.
If each item image yields $R+1$ patch tokens (including a \texttt{[CLS]} token), ten items produce $10 \times (R+1)$ image tokens.
Because self-attention scales quadratically with sequence length, this \emph{token explosion} can exceed the LLM context window and memory budget.
Moreover, it makes training unwieldy, causing the model to struggle with learning useful representations, and thereby dropping recommendation performance.
For instance, LLaVA-v1.6 sets $R=576$ for each of $5$ sub-images and uses a $4,096$ token context length, so handling ten images at once is infeasible under these constraints.

\section{Reddit-Amazon Dataset}
\label{sec:dataset}
Although many conversational recommendation datasets have been introduced, most rely on crowd-sourced or synthetic dialogues (e.g., ReDial \cite{LiKSMCP18}, GoRecDial \cite{KangBSCBW19}, and TG-ReDial \cite{ZhouZZWW20}).
Crowd workers often lack specific personal interest, creating artificial queries and interactions \cite{HeXJSLFMKM23}, while fully synthetic dialogues are constructed from user behaviors or other data and may fail to capture the spontaneity and depth of real conversations.
Datasets reflecting authentic user interactions (e.g., E-ConvRec \cite{JiaLW0XLSCP022} and Reddit-Movie \cite{HeXJSLFMKM23}) tend to focus on domains such as movies or electronics, where visual attributes (e.g., color, style, or design) play a less prominent role.
U-Need~\cite{00010DFWFCZCLC23} covers categories like fashion and beauty, but does not include visual sources.
These constraints of existing datasets limit their applicability to realistic visually-aware conversational recommendation.

To address these gaps, we collect real user data from Reddit via \emph{pushshift.io}\footnote{\url{https://pushshift.io}} in three visually oriented domains: beauty, fashion, and home.
We first identify relevant posts and comments using GPT-3.5-turbo~\cite{openai2022chatgpt} and a filtering prompt (Table~\ref{tab:filtering_prompt} in Appendix~\ref{app:prompt}), then link the mentioned items to the Amazon Reviews 2023 dataset~\cite{abs-2403-03952}.
The resulting \emph{Reddit–Amazon} dataset is divided into three subsets (beauty, fashion, and home), each containing genuine user dialogues, ground-truth recommended items, and corresponding item features including images.
Figure~\ref{fig:data_illustration} provides an illustration and Table~\ref{tab:dataset_stats} presents key statistics for \emph{Reddit–Amazon}, offering a realistic benchmark for evaluating visually-aware conversational recommendation.
We also compare \emph{Reddit-Amazon} dataset with existing conversational recommendation datasets in Table~\ref{tab:convrec_datasets} (Appendix~\ref{app:dataset_convrec}).

\begin{table}[t]
    \centering
    \caption{
        Summary of \emph{Reddit-Amazon} dataset. The \emph{Reddit-Amazon} dataset consists of three sub-categories based on the type of recommended items: \emph{beauty}, \emph{fashion}, and \emph{home}.
    }
    \begin{tabular}{lcccc}
        \toprule
        \textbf{Dataset} & \textbf{\# Conv.} & \textbf{\# Turns} & \textbf{\# Items} & \textbf{\# Images} \\
        \midrule
        \textit{Beauty} & 7,672 & 22,966 & 5,433 & 28,082\\
        \textit{Fashion} & 8,039 & 21,831 & 6,716 & 31,162\\
        \textit{Home} & 3,701 & 6,675 & 3,077 & 18,505\\
        \bottomrule
    \end{tabular}
\label{tab:dataset_stats}
\end{table}

% \red{All beauty: 8,424 $\rightarrow$ 6,480 after recommendation filtering using ChatGPT-3.5-t.
% Amazon fashion: 10,314 $\rightarrow$ 7,460 after recommendation filtering using ChatGPT-3.5-t.

% Unique items: 12,091.
% Images: 56,909 (11 fail cases due to the url error are not included).

% All beauty: 6,884 $\rightarrow$ 6,340 after noise filtering with threshold 10.
% After split, train: 5,072, valid: 634, and test: 634 (8:1:1).
% Amazon fashion: 8,370 $\rightarrow$ 7,733 after noise filtering with threshold 10.
% After split, train: 6,186, valid: 773, and test: 774 (8:1:1).}

% We have 7,917, 8,320, and 3,701 conversations (total 19,938) for All Beauty, Amazon Fashion, and Amazon Home, respectively, after recommendation filtering using ChatGPT-3.5-t.

% Unique items: 15,999.
% Images: 81,061 (10 fail cases due to url errors).
% All Beauty (instances): 7,672 $\rightarrow$ 7,825 after noise filtering (6,142, 758, 772).
% All Beauty (unique items): 5,565.
% Amazon Fashion (instances): 8,039 $\rightarrow$ 8,265 after noise filtering (6,416, 797, 826).
% Amazon Fashion (unique items): 6,627.
% Amazon Home (instances): 3,701 $\rightarrow$ 3,701 after noise filtering (2,961, 368, 372).
% Amazon Home (unique items): 2,971.

% Fake images: 243,213.

% Item description data (title dependent generation).

\begin{figure*}[t]
	\centering
	\includegraphics[width=1.0\linewidth]{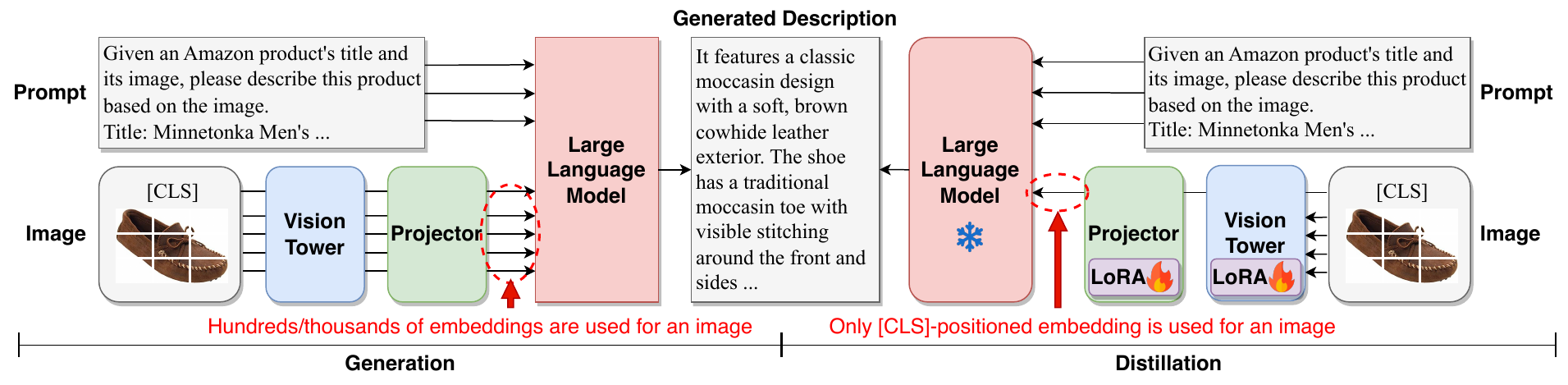}
	% \caption{
 %        Illustration of visual knowledge self-distillation.
 %        We compress each sub-image from hundreds/thousands of tokens (577 for each sub-image in LLaVA-v1.6) into a single \texttt{[CLS]}-positioned embedding.
	% }
    \caption{
        Illustration of \emph{visual knowledge self-distillation}. Generation process (left): The vision tower and projector encode each sub-image into hundreds of patch embeddings (577 for each sub-image in LLaVA-v1.6), which are passed to the large language model (LLM) alongside a textual prompt. The LLM then produces a detailed product description focusing on visual features. Distillation process (right): We freeze the LLM and train only the vision tower and projector (via LoRA) to condense each sub-image into a single \texttt{[CLS]}-positioned embedding, yet still generate the same descriptive text. This reduces the token count from thousands to a handful, minimizing computational overhead while retaining essential visual information.
    }
\label{fig:self_distillation}
\end{figure*}

\section{Visually-Aware Conversational Recommendation}
\label{sec:proposed}

In this section, we propose \method, a two-stage framework that addresses token explosion in \emph{visually-aware conversational recommendation} by compressing images into fewer tokens and then fine-tuning a large vision-language model (VLM) for accurate recommendation.

\subsection{Framework Overview}
\label{subsec:overview}

We build upon LLaVA-v1.6, which is recognized for its adaptability to a broad range of downstream tasks and demonstrates strong performance in multi-image scenarios~\cite{liu2024llavanext}.
LLaVA-v1.6 encodes each image by splitting it into 5 sub-images, each producing $577$ tokens for a total of $2,885$ tokens per image.
Analyzing multiple items in a single query can quickly surpass the 4,096 token context limit, making naive end-to-end training unstable and limiting recommendation accuracy (Section~\ref{subsec:lvms}).
To resolve these issues, \method employs two key stages:

\begin{itemize}[leftmargin=*]
    \item \textbf{Visual knowledge self-distillation (Section~\ref{subsec:distillation}).} We freeze the parameters of the large language model and train only the vision module (vision tower and projector) to condense each item’s 5 sub-images $(5 \times 577$ tokens) into a small set of \texttt{[CLS]} embeddings. A task-oriented prompt, asking for visually relevant attributes, guides this distillation.
    \item \textbf{Recommendation prompt tuning (Section~\ref{subsec:prompt_tuning}).} We then freeze the distilled vision module and fine-tune the large language model. Given the compressed image embeddings and textual context, the model predicts the correct item ID between $10$ candidates, thus preventing the risk of generating nonexistent items.
\end{itemize}

\subsection{Visual Knowledge Self-Distillation}
\label{subsec:distillation}

We aim to produce the same image description with far fewer tokens.
Figure~\ref{fig:self_distillation} illustrates how we distill vision knowledge into fewer tokens in a self-distillation manner.
Initially, the VLM sees each image with all sub-image tokens (i.e., $2,885$ tokens if $5$ sub-images are each mapped to $577$ tokens) and generates a detailed description.
We then distill this capability so that the large language model (LLM) can regenerate the same description from a single \texttt{[CLS]} embedding per sub-image.
This reduces the model’s reliance on thousands of tokens, preventing overflow of the context window in candidate-based recommendations where multiple items appear in a single query.

\para{Generation.}
In the generation process, the VLM (with parameters $\Omega_{\mathrm{LM}} + \Omega_{\mathrm{vision}}$) is given an instruction to generate the description of an item image, along with the entire set of sub-image tokens for $\mathit{image}_i$;
the prompt used for generation is detailed in Table~\ref{tab:description_prompt} (Appendix~\ref{app:prompt}).
The model freely attends to all these tokens, eventually producing a textual description $D_i$ (e.g., specifying color, material, or design).
This step shows the ability of the VLM to generate visually rich output, but at the cost of processing thousands of tokens per image.
We provide examples of these generated descriptions in Appendix~\ref{app:generation_examples}.

\para{Distillation.}
Then, we freeze $\Omega_{\mathrm{LM}}$ (the parameters of LLM) and keep only the vision-side parameters $\Omega_{\mathrm{vision}}$ trainable to focus on distilling only visual capability.
We utilize only the \texttt{[CLS]}-positioned embedding for each sub-image that must alone suffice to generate $D_i$.
Formally, if $\mathit{image}_i$ is divided into 5 sub-images $\{I_{i,r}\}_{r=1}^{5}$, each sub-image $I_{i,r}$ yields $\texttt{Tok}_\mathrm{vision}(I_{i,r}) \in \mathbb{R}^{577\times d}$ in the original model.
Rather than passing all $577$ tokens to the LLM, we extract the \texttt{[CLS]}-positioned embedding which we denote as $\mathbf{cls}_{i,r}\,\in\mathbb{R}^d$.
Concatenating $\{\mathbf{cls}_{i,r}\}$ across all $5$ sub-images yields $5$ embeddings per entire image.
We replace the original sub-image tokens in the input sequence with these 5 \texttt{[CLS]}-positioned embeddings and then prompt the frozen LLM to generate $D_i$.

\begin{figure}[t]
	\centering
	\includegraphics[width=0.6\linewidth]{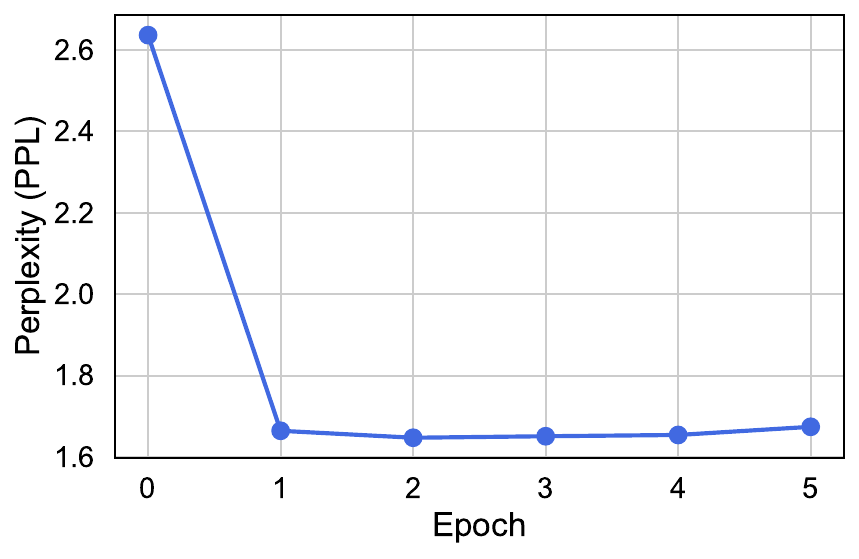}
	\caption{
        The validation perplexity reaches a plateau after 1--2 epochs.
	}
\label{fig:perplexity}
\end{figure}

\para{Training objective.}
We train the vision module so that the LLM’s output, given only $5$ \texttt{[CLS]}-positioned embeddings per image, reproduces the same description $D_i$ that was previously generated from all sub-image tokens.
This drives each $\mathbf{cls}_{i,r}$ to encapsulate the most crucial visual details. Concretely, we optimize:
\begin{equation}
\min_{\Omega_{\mathrm{vision}}}
\sum_i
- \log 
P_{\Omega_{\mathrm{LM}} + \Omega_{\mathrm{vision}}}
\bigl(D_i \mid \mathcal{T}_\mathrm{desc}, \{\mathbf{cls}_{i,r}\}_{r=1}^{5}\bigr),
\end{equation}
where $\mathcal{T}_\mathrm{desc}$ is the textual prompt asking for visually relevant attributes.
We adopt LoRA~\cite{HuSWALWWC22} in the vision tower and the projector to keep the overhead parameter minimal.
Figure~\ref{fig:perplexity} shows the perplexity (\textsc{PPL}) on a held-out set of 512 images and their generated descriptions; it typically converges after $1$--$2$ epochs, showing the effectiveness of our focus on a small set of trainable parameters.
We select the checkpoint with the lowest \textsc{PPL} ($2$ in our case) for the subsequent recommendation tasks.

After this self-distillation, each image is effectively represented by only $5$ \texttt{[CLS]}-positioned embeddings rather than thousands of tokens, yet the frozen LLM can still reconstruct $D_i$.
This preserves descriptive power with minimal visual tokens, laying the groundwork for candidate-based conversational recommendation, where multiple images may appear in a single query, without exceeding the model context.

\subsection{Recommendation Prompt Tuning}
\label{subsec:prompt_tuning}

Once the vision parameters $\Omega_{\mathrm{vision}}$ are obtained after the visual knowledge self-distillation (Section~\ref{subsec:distillation}), we focus on generating accurate recommendations for a given prompt.
In turn $k$ of a conversation, the model receives a dialogue context $\mathbf{C} = \{s_t\}_{t=1}^{k-1}$ between the seeker and the recommender.
A retrieval module (e.g., SBERT~\cite{ReimersG19}) then presents $10$ candidate items $\{i_1,\dots,i_{10}\}$. 
Each item $i_j$ is represented by a textual title $\mathit{title}_{i_j}$ and $5$ distilled \texttt{[CLS]}-positioned embeddings $\{\mathbf{cls}_{i_j,r}\}_{r=1}^5$.
We make sure that exactly one candidate $i^*$ is correct for each training example.
In other words, if there are multiple ground-truths, we split the example into multiple training instances.
Besides, if $i^*$ is initially absent in the candidates, we swap out one negative candidate for $i^*$ to ensure precisely one ground-truth item.
We also randomly shuffle the candidate items to avoid any bias in their order.

\begin{figure}[t]
	\centering
	\includegraphics[width=1.0\linewidth]{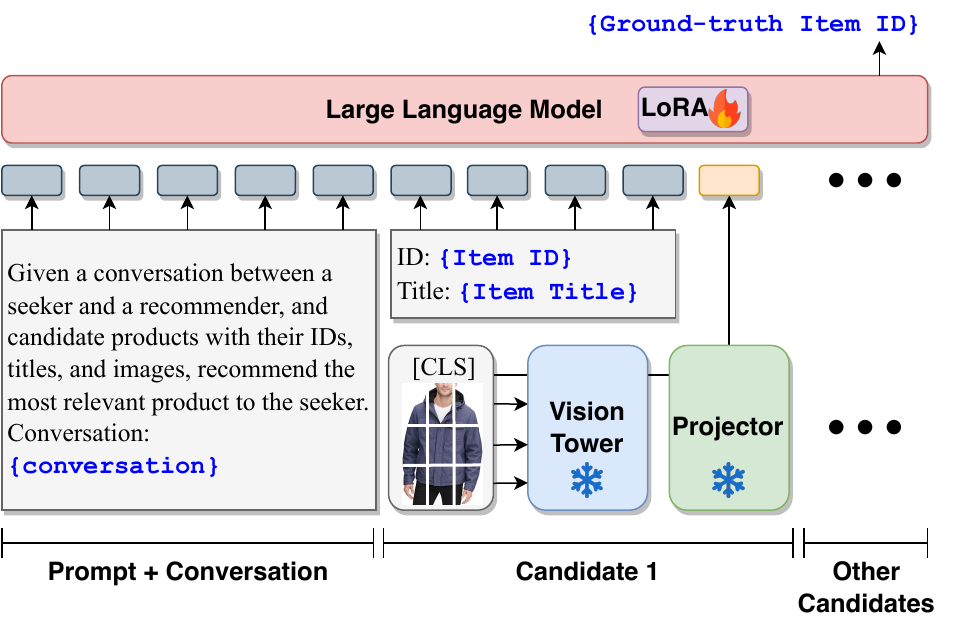}
	\caption{
        Illustration of recommendation prompt tuning.
        We integrate the compressed embeddings with conversation and item IDs/titles for recommendation.
        We then train the large language model (LLM) using LoRA while fixing the parameters of vision tower and projector.
	}
\label{fig:prompt_tuning}
\end{figure}

\para{Prompt setting.}
Figure~\ref{fig:prompt_tuning} illustrates how we feed textual information and each candidate's compressed image embeddings into the LLM.
For each item \(i_j\), we concatenate its ID $\mathit{id}_{i_j}$ and textual title $\mathit{title}_{i_j}$ with its $5$ \texttt{[CLS]}-positioned embeddings $\{\mathbf{cls}_{i_j,r}\}_{r=1}^5$ as follows:
\begin{equation}
\mathbf{X}_{i_j} \;=\;
\mathrm{Concat}\Bigl(\mathit{id}_{i_j}, \mathit{title}_{i_j},\;\mathbf{cls}_{i_j,1},\dots,\mathbf{cls}_{i_j,5}\Bigr).
\end{equation}
We then supply \(\{\mathbf{X}_{i_j}\}_{j=1}^{10}\) and the tokenized conversation context with a task description prompt $\mathcal{T}_\mathrm{conv}$ to the LLM, which is instructed to output a short item ID $\mathit{id}_{i_j}$ (e.g., 10 characters) for $i^*$. 
This ID-based approach normalizes the system's output, avoiding hallucinated item names that do not appear in the entire set of items \(\mathcal{I}\).

\para{Training objective.}
We train the LLM so that it generates the exact ID of the ground-truth item assuming that the vision module is already trained to extract crucial visual features.
Each instance contains exactly one ground-truth item ($i^*$) and nine negatives, restricting the output to a concise ID.
Let $\mathit{id}_{i^*}$ denote the ID of the ground-truth item.
We aim to generate $\mathit{id}_{i^*}$ from the dialogue context $\mathcal{T}_\mathrm{conv}$ and the representation of each candidate $\{\mathbf{X}_{i_j}\}_{j=1}^{10}$.
Specifically, we optimize:
\begin{equation}
\min_{\Omega_{\mathrm{LM}}}
\sum_{(\mathcal{T}_\mathrm{conv},\,\mathcal{I}_\mathrm{cand})}
- \log
P_{\Omega_{\mathrm{LM}} + \Omega_{\mathrm{vision}}}
\Bigl(\mathit{ID}_{i^*}
\;\big|\;
\mathcal{T}_\mathrm{conv}, \{\mathbf{X}_{i_j}\}_{j=1}^{10}\Bigr),
\end{equation}
where $\mathcal{I}_\mathrm{cand}$ denotes the candidate items in an example.
We adapt LoRA exclusively to the LLM's parameters $\Omega_{\mathrm{LM}}$ while fixing the distilled vision module $\Omega_{\mathrm{vision}}$.

In inference, the core visual features of each candidate item are compactly encoded as $5$ \texttt{[CLS]}-positioned embeddings, while the LLM receives the dialogue context, item IDs, and titles.
Drawing on these visual and textual inputs, the model determines which candidate is best aligned with the user's preferences and outputs the corresponding item ID. 
Although the vision module supplies only a handful of tokens per image, it preserves sufficient detail for domains with high visual complexity as shown in Figure~\ref{fig:perplexity}.
Simultaneously, the LLM remains within its context limits and can robustly integrate multi-item information. 
Thus, \method achieves accurate recommendations without incurring token overflow, making it suitable for practical candidate-based pipelines.

\section{Experiments}
\label{sec:experiment}

\subsection{Experimental Setup}
\label{subsec:experimental_setup}

\para{Datasets.}
We use the newly constructed \emph{Reddit-Amazon} datasets, which consist of three different domains: beauty, fashion, and home.
Refer to Section~\ref{sec:dataset} for details and Table~\ref{tab:dataset_stats} for a summary.
For each domain, we divide conversations into training, validation, and test sets in an 8:1:1 ratio.
For multiple ground-truth items, we construct the conversation as multiple examples to make each example have a single ground-truth.

\para{Baselines.}
We compare our proposed method with two categories of baselines: retrieval-based and generative methods.
Although there exist earlier knowledge graph-based conversational recommendation methods (e.g., KBRD~\cite{ChenLZDCYT19}, KGSF~\cite{ZhouZBZWY20}, and UniCRS~\cite{WangZWZ22}), they rely heavily on knowledge graphs and have mainly been evaluated on movie domains.
Generalizing such approaches to diverse product categories (beauty, fashion, home) is non-trivial.
Moreover, prior studies indicate that zero-shot large language models (LLMs) often surpass such knowledge graph-based conversational recommender systems on Reddit dialogues~\cite{HeXJSLFMKM23}. 
Thus, we focus on retrieval-based and generative baselines that can flexibly handle our \emph{Reddit-Amazon} dataset.

\noindent
\textbf{(1) Retrieval methods.}
All following methods except BM25 leverage pre-trained models to encode the conversation text and item titles, then rank items by cosine similarity.
% BM25 is a classic lexical matcher.
The top-ranked item is returned as the recommendation.
\begin{itemize}
    \item \textbf{BM25}~\cite{RobertsonZ09}: It scores titles by token overlap with the conversation text.
    \item \textbf{SBERT}~\cite{ReimersG19}: It generates sentence embeddings via a siamese BERT-based approach.
    \item \textbf{RoBERTa\textsubscript{large}}~\cite{abs-1907-11692}: It replicates BERT with careful hyperparameter tuning and larger training data.
    \item \textbf{SimCSE\textsubscript{large}}~\cite{GaoYC21}: It employs a contrastive learning objective for sentence embeddings.
    \item \textbf{BLaIR\textsubscript{base}}~\cite{abs-2403-03952}: It is a specialized sentence embedding model for recommendation, trained on a large-scale Amazon review dataset to capture item-text correlations.
    \item \textbf{OpenAI-emb\textsubscript{large}}\footnote{\url{https://platform.openai.com/docs/guides/embeddings}}: It is a proprietary OpenAI\footnote{\url{https://openai.com}} model, one of the most powerful encoders for many complex textual contexts.
\end{itemize}

\noindent
\textbf{(2) Generative methods.}
Each method first retrieves the top-10 candidates (using one of the above retrieval models), then employs a shared prompt template to generate a single recommendation from those candidates.
\begin{itemize}
    \item \textbf{Vicuna-v1.5}~\cite{vicuna2023}: It is a fine-tuned version of LLaMA~\cite{LiuLWL23a} on user-shared conversations from ShareGPT.
    \item \textbf{LLaVA-v1.5}~\cite{liu2023improvedllava}: It extends Vicuna for vision-language tasks.
    \item \textbf{LLaVA-v1.6}~\cite{liu2024llavanext}: It is fine-tuned for multi-image tasks with using Mistral~\cite{abs-2310-06825} as an LLM.
    \item \textbf{GPT-3.5-turbo}~\cite{openai2022chatgpt}, \textbf{GPT-4o-mini}~\cite{abs-2410-21276}, and \textbf{GPT-4o}: These are OpenAI proprietary models with powerful abilities to solve complex tasks in a zero-shot environment~\cite{abs-2303-12712}.
\end{itemize}

\begin{table}[t]
	\setlength\tabcolsep{4pt}
    \small
    \centering
    \caption{
        Performance comparison of \method with open-source baselines.
        \colorbox{gray!10}{Gray colored} methods require a candidate retrieval to generate answers.
        We use SBERT as the common retrieval since it shows the best performance among the open-source retrievals.
        Bold and underline indicate the best and the second-best, respectively.
    }
    \begin{tabular}{l|cc|cc|cc}
        \toprule
        \multirow{2}{*}{\textbf{Method}} & \multicolumn{2}{c}{\textit{\textbf{Beauty}}} & \multicolumn{2}{c}{\textit{\textbf{Fashion}}} & \multicolumn{2}{c}{\textit{\textbf{Home}}}\\
        \cmidrule(lr){2-3} \cmidrule(lr){4-5} \cmidrule(lr){6-7}
        & \textbf{HR@1} & \textbf{VR} & \textbf{HR@1} & \textbf{VR} & \textbf{HR@1} & \textbf{VR}\\
        \midrule
        \multicolumn{7}{c}{\textit{Retrieval Baselines (item title)}}\\
        BM25 & 0.0169 & - & 0.0140 & - & 0.0479 & -\\
        SBERT & 0.0551 & - & 0.0681 & - & \underline{0.2166} & -\\
        RoBERTa\textsubscript{large} & 0.0640 & - & 0.0631 & - & 0.1814 & -\\
        SimCSE\textsubscript{large} & 0.0326 & - & 0.0301 & - & 0.0957 & -\\
        BLaIR\textsubscript{base} & 0.0371 & - & 0.0441 & - & 0.1335 & -\\
        \midrule
        \rowcolor{gray!10}\multicolumn{7}{c}{\textit{Generative Baselines (item title)} + SBERT}\\
        \rowcolor{gray!10}Vicuna-v1.5 & 0.0533 & 0.9870 & 0.0481 & 0.9903 & 0.1184 & 1.0000\\
        \rowcolor{gray!10}LLaVA-v1.5 & 0.0476 & 0.9896 & 0.0441 & 0.9855 & 0.0932 & 1.0000\\
        \rowcolor{gray!10}LLaVA-v1.6 & \underline{0.0770} & 0.9870 & \underline{0.0827} & 0.9867 & 0.2030 & 0.9919\\
        \midrule
        \rowcolor{gray!10}\multicolumn{7}{c}{\textit{Generative Baselines (item title and image)} + SBERT}\\
        \rowcolor{gray!10}LLaVA-v1.5 & 0.0000 & 0.0000 & 0.0000 & 0.0000 & 0.0000 & 0.0000 \\
        \rowcolor{gray!10}LLaVA-v1.6 & 0.0584 & 0.9741 & 0.0459 & 0.9843 & 0.1089 & 0.9919\\
        \midrule
        \rowcolor{gray!10}\multicolumn{7}{c}{\textit{Proposed Method (item title and image)} + SBERT} \\
        \rowcolor{gray!10}\textbf{\method (ours)} & \textbf{0.1187} & 0.9702 & \textbf{0.1232} & 0.9298 & \textbf{0.3197} & 0.9892\\
        \rowcolor{cyan!20}\textbf{Improvement} & \textbf{+54.2\%} & - & \textbf{+49.0\%} & - & \textbf{+47.6\%} & -\\
        \bottomrule
    \end{tabular}
    \label{tab:comparison_open_sbert}
\end{table}

\para{Implementation details.}
We conduct all experiments on a single \emph{NVIDIA A100 40GB} GPU.
We obtain open-source pretrained models (SBERT, RoBERTa, SimCSE, BLaIR, Vicuna, and LLaVA) from the official HuggingFace\footnote{\url{https://huggingface.co}} repositories, while GPT embeddings, GPT-3.5-turbo, GPT-4o-mini, and GPT-4o are accessed via the OpenAI API.
All open-source generative methods use the 7B parameter scale (i.e., Vicuna-v1.5-7B, LLaVA-v1.5-7B, LLaVA-v1.6-7B, and \method-7B (ours)) to enable both training and evaluation on a single GPU.
For generative approaches (including \method), we adopt a candidate-based pipeline: 
(1) retrieve the top-$10$ items for each conversation, then (2) generate a single recommendation via an LLM prompt.
We specifically employ SBERT and OpenAI-emb\textsubscript{large} as retrieval methods: SBERT gives the best performance among open-source retrieval baselines, whereas OpenAI-emb\textsubscript{large} performs best when proprietary methods are considered.
Further hyperparameter and training details for our method appear in Appendix~\ref{app:hyperparams}.

\para{Evaluation Metrics.}
Our primary metric is \emph{HitRatio@1 (HR@1)}, which measures the performance of the recommendation by computing the proportion of conversations where the prediction is equal to the ground-truth item.
For generative methods, we additionally report a \emph{ValidRatio (VR)} following a previous work~\cite{LiaoL0WYW024}, defined as the fraction of generated responses that match one of the candidate items; it measures the model’s adherence to the prompt instructions.

\begin{table}[t]
    \setlength\tabcolsep{4.2pt}
    \small
    \centering
    \caption{
        Performance comparison of \method with proprietary baselines.
        We use SBERT as a candidate retrieval.
    }
    \begin{tabular}{l|cc|cc|cc}
        \toprule
        \multirow{2}{*}{\textbf{Method}} & \multicolumn{2}{c}{\textit{\textbf{Beauty}}} & \multicolumn{2}{c}{\textit{\textbf{Fashion}}} & \multicolumn{2}{c}{\textit{\textbf{Home}}}\\
        \cmidrule(lr){2-3} \cmidrule(lr){4-5} \cmidrule(lr){6-7}
        & \textbf{HR@1} & \textbf{VR} & \textbf{HR@1} & \textbf{VR} & \textbf{HR@1} & \textbf{VR}\\
        \midrule
        \multicolumn{7}{c}{\textit{Generative Baselines (item title)} + SBERT}\\
        GPT-3.5-turbo & 0.0968 & 0.9935 & 0.0977 & 0.9903 & 0.2343 & 1.0000\\
        GPT-4o-mini & 0.1213 & 1.0000 & 0.1160 & 0.9927 & 0.3258 & 0.9973\\
        GPT-4o & 0.1271 & 0.9987 & 0.1278 & 0.9976 & 0.3350 & 1.0000\\
        % GPT-4o (ZS-BM25) & 0.1077 & - & 0.1205 & - & 0.2177 & -\\
        % GPT-4o (ZS-ED) & 0.0000 & - & 0.0000 & - & 0.0000 & -\\
        \midrule
        \multicolumn{7}{c}{\textit{Generative Baselines (item title and image)} + SBERT}\\
        GPT-4o-mini & 0.1081 & 0.9974 & 0.1098 & 0.9927 & 0.2861 & 0.9946\\
        GPT-4o & 0.1160 & 0.9974 & 0.1231 & 0.9939 & 0.3308 & 0.9973\\
        \midrule
        \multicolumn{7}{c}{\textit{Proposed Method (item title and image)} + SBERT} \\
        \textbf{\method (ours)} & 0.1187 & 0.9702 & 0.1232 & 0.9298 & 0.3197 & 0.9892\\
        \bottomrule
    \end{tabular}
    \label{tab:comparison_close_sbert}
\end{table}

\subsection{Overall Performance}
\label{subsec:performance}

We compare our method with two different categories of baselines: open-source baselines and proprietary baselines.
For generative baselines and our method, we rely on a candidate-based recommendation setup with $10$ items, using SBERT~\cite{ReimersG19} which shows the highest performance between retrievals.
Generative baselines are tested under two configurations: using only item titles (\textit{item title}) or using both titles and images (\textit{item title and image}). 

\para{Comparison with open-source baselines.}
Table~\ref{tab:comparison_open_sbert} shows the comparison of \method with open-source baselines. \emph{Retrieval-based} baselines (BM25, SBERT, RoBERTa\textsubscript{large}, SimCSE\textsubscript{large}, and BLaIR\textsubscript{base}) rank items by textual similarity only.
Their effectiveness varies by domain.
For example, in Home, SBERT achieves $HR@1=0.2166$, which exceeds LLaVA-v1.6 ($HR@1=0.2030$) when both rely on titles alone.
In contrast, in Beauty and Fashion, LLaVA-v1.6 with titles outperforms these retrieval methods.
We also observe that incorporating images without specialized handling can degrade the performance of generative models, which is consistent with a previous work~\cite{abs-2402-08670}.
LLaVA-v1.6 shows reduced accuracy in most settings when images are used, and LLaVA-v1.5 fails in multi-image candidate scenarios dropping VR to zeros on all domains;
LLaVA-v1.6 is trained under multi-image tasks, while LLaVA-v1.5 is not.
In contrast, \method achieves the highest HR@1 in all domains, up to 54.2\% higher performance than the second-best.
Furthermore, \method maintains a high VR, indicating consistency in generating valid outputs.
These results suggest that compressing each item image into several \texttt{[CLS]}-positioned embeddings effectively preserves critical visual details while preventing token explosion, thus yielding greater recommendation accuracy in visually intensive domains.

\para{Comparison with proprietary baselines.}
Table~\ref{tab:comparison_close_sbert} compares \method with proprietary baselines.
They also rely on SBERT for candidate retrieval and then generate an item ID using textual titles or both titles and images.
GPT-4o generally achieves the highest HR@1 in all domains, showing its powerful zero-shot performance in complex downstream tasks.
GPT-4o-mini and GPT-4o both lose accuracy when images are included, indicating that multi-image processing can be challenging in complex conversational recommendations.
Our method obtains performance on par with GPT-4o-mini or GPT-4o, and exceeds GPT-3.5-turbo in all domains, despite relying on 7B-parameters.
These results show that distilled vision embeddings coupled with an LLM can handle multi-item visual contexts as effectively as or better than larger commercial systems while mitigating token explosion.
As a result, \method provides a strong balance of visual representation efficiency and recommendation accuracy in domains where product appearance strongly influences user choice.

We also compare performance using OpenAI-emb\textsubscript{large}, which is known to deliver strong performance in various tasks, as a retrieval in Appendix~\ref{app:experiments}.
The results show that the overall trends remain consistent with those obtained in Tables~\ref{tab:comparison_open_sbert} and~\ref{tab:comparison_close_sbert}.

\begin{table}[t]
    \setlength\tabcolsep{2pt}
    \small
    \centering
    \caption{
        Ablation study. SBERT is used for retrieval. \emph{o.o.m.} indicates out-of-memory errors that prevent running under our single-GPU setup.
    }
    \begin{tabular}{l|cc|cc|cc}
        \toprule
        \multirow{2}{*}{\textbf{Method}} & \multicolumn{2}{c}{\textit{\textbf{Beauty}}} & \multicolumn{2}{c}{\textit{\textbf{Fashion}}} & \multicolumn{2}{c}{\textit{\textbf{Home}}}\\
        \cmidrule(lr){2-3} \cmidrule(lr){4-5} \cmidrule(lr){6-7}
        & \textbf{HR@1} & \textbf{VR} & \textbf{HR@1} & \textbf{VR} & \textbf{HR@1} & \textbf{VR}\\
        \midrule
        Entire tokens ($5\times577$) & 0.0256 & 0.9456 & \multicolumn{2}{c}{\emph{o.o.m.}} & \multicolumn{2}{c}{\emph{o.o.m.}}\\
        \emph{w/o} images & \underline{0.0972} & 0.9767 & 0.1022 & 0.9358 & \underline{0.2944} & 0.9946\\
        \emph{w/o} self-distillation & 0.0842 & 0.9793 & \underline{0.1084} & 0.9649 & 0.2861 & 0.9973\\
        \midrule
        \textbf{\method (ours)} & \textbf{0.1187} & 0.9702 & \textbf{0.1232} & 0.9298 & \textbf{0.3197} & 0.9892\\
        \bottomrule
    \end{tabular}
    \label{tab:ablation}
\end{table}

\subsection{Ablation Study}
\label{subsec:ablation}
Table~\ref{tab:ablation} compares \method with three variants, each omitting a main component of our framework.
\emph{Entire tokens} ($5\times577$) applies recommendation prompt tuning to LLaVA-v1.6 directly, processing all $2,885$ tokens per image.
In the beauty domain, this configuration requires about one week for a single epoch on two A100 GPUs and yields low accuracy (HR@1=0.0256).
Given these resource constraints, we cannot run the fashion and home configurations on our single-GPU setup, so we mark them as \emph{o.o.m.}.
In \emph{w/o images}, the model eliminates visual input, relying solely on item titles, and achieved higher accuracy than \emph{Entire tokens} (e.g., HR@1 = 0.0972 in beauty).
However, this shows a lower HR@1 than \method in all domains, since it ignores visual characteristics.
The \emph{w/o self-distillation} preserves \texttt{[CLS]} extraction per sub-image but bypasses our visual knowledge distillation stage, further improving over \emph{Entire tokens} but still underperforming \method. 
In contrast, \method incorporates both visual knowledge self-distillation and recommendation prompt tuning, reducing each image to $5$ \texttt{[CLS]}-positioned embeddings with minimal loss of visual detail.
\method attains the best HR@1 and strong VR across all domains, demonstrating token compression via self-distillation is crucial for balancing memory usage, training time, and accuracy in visually-aware conversational recommendations.

\subsection{Further Analysis}
\label{subsec:further}

\subsubsection{Separate dataset vs. combined dataset}
\label{subsubsec:separate_combined}
Table~\ref{tab:separate_combined} compares two training strategies for \method.
In \method-\emph{separate} training, we learn a different model in each domain (beauty, fashion, and home) and then evaluate it in the same domain.
In \method-\emph{combined} training, we merge all three domains’ data into a single training set and test on each domain separately.
The results show that separate training yields slightly higher HR@1 than the combined approach.
We hypothesize that combining data from multiple domains does not confer immediate benefits because the recommended items do not overlap between beauty, fashion, and home, and each domain’s user preference is distinct.
These factors limit cross-domain information sharing.
Another consideration is that we fixed a single set of hyperparameters for the merged dataset, which may not optimally fit the specialized characteristics of each domain.
However, our dataset remains small in size compared to many industrial-scale corpora.
It is plausible that with more extensive data and broader conversational diversity, the model might learn cross-domain representations that improve performance.
Hence, future studies could explore whether larger-scale domain mixtures further enhance performance in visually-aware conversational recommendations.

\begin{table}[t]
    \setlength\tabcolsep{3.5pt}
    \small
    \centering
    \caption{
        Separate dataset vs. combined dataset.
    }
    \begin{tabular}{l|cc|cc|cc}
        \toprule
        \multirow{2}{*}{\textbf{Method}} & \multicolumn{2}{c}{\textit{\textbf{Beauty}}} & \multicolumn{2}{c}{\textit{\textbf{Fashion}}} & \multicolumn{2}{c}{\textit{\textbf{Home}}}\\
        \cmidrule(lr){2-3} \cmidrule(lr){4-5} \cmidrule(lr){6-7}
        & \textbf{HR@1} & \textbf{VR} & \textbf{HR@1} & \textbf{VR} & \textbf{HR@1} & \textbf{VR}\\
        \midrule
        \textbf{\method}-\emph{separate} & \textbf{0.1187} & 0.9702 & \textbf{0.1232} & 0.9298 & \textbf{0.3197} & 0.9892\\
        \textbf{\method}-\emph{combined} & 0.1021 & 0.9585 & 0.1220 & 0.9673 & 0.3141 & 0.9651\\
        \bottomrule
    \end{tabular}
    \label{tab:separate_combined}
\end{table}

\subsubsection{Case study}
\label{subsubsec:case_study}
\begin{figure}[t]
	\centering
	\includegraphics[width=1.0\linewidth]{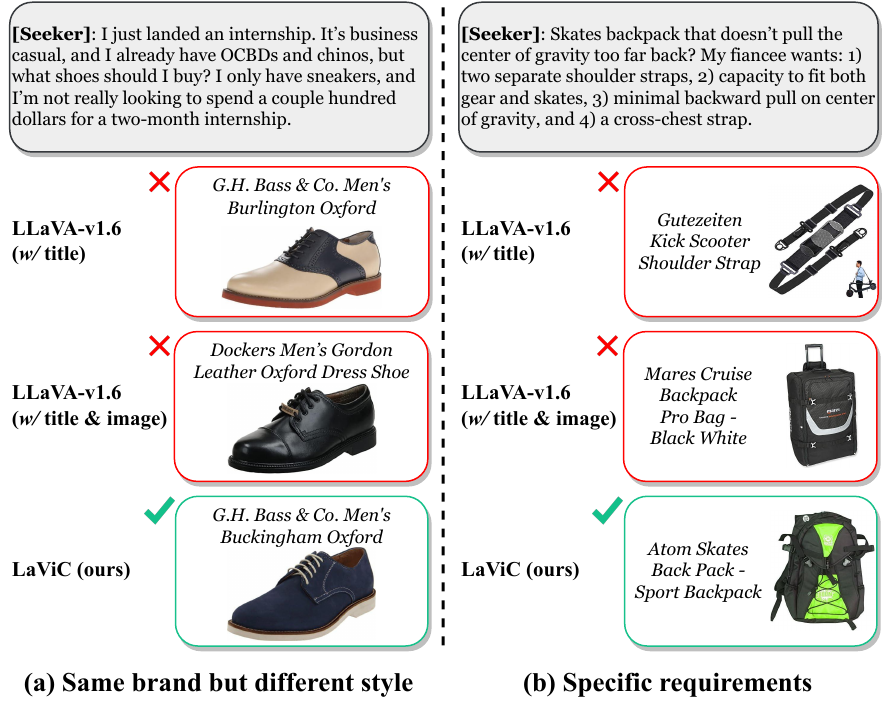}
    \caption{
        Two cases comparing \method with LLaVA-v1.6. Both cases are selected from \emph{fashion} domain. The \cmark\xspace indicates a correct recommendation, while the \xmark\xspace denotes an incorrect one. (a) \method identifies a more casual ``G.H. Bass \& Co. Men’s Buckingham Oxford'', whereas LLaVA-v1.6 either suggests a more formal shoe under the same brand or a different brand entirely. (b) \method recommends a backpack that meets the user’s specific strap requirements (two separate shoulder straps and a cross-chest strap), which LLaVA-v1.6 (\emph{w/} title \& image) fails to satisfy, even with image input.
    }

\label{fig:case_study}
\end{figure}

Figure~\ref{fig:case_study} illustrates two cases where \method outperforms two variants of LLaVA-v1.6: one using only titles (\emph{w/} title) and the other combining titles and images (\emph{w/} title \& image).
We use OpenAI-emb\textsubscript{large} as the retriever to select $10$ relevant candidates for each scenario, showing how well each method handles subtle differences among highly relevant items.
See Table~\ref{tab:comparison_open_openai} in Appendix~\ref{app:experiments} for a quantitative comparison.

\para{(a) Same brand but different style.}
The seeker wants a shoe for business-casual wear, and the ground-truth item is from ``G.H. Bass \& Co. Men’s Buckingham Oxford'' with a casual design.
LLaVA-v1.6 (\emph{w/} title) recommends a product of the same brand, but with a more formal look.
LLaVA-v1.6 (\emph{w/} title \& image) fails to match the brand and casual style the seeker prefers.
In contrast, \method selects an item from ``G.H. Bass \& Co.'' that better matches the intended style, relying on subtle visual features missing from the textual title.

\para{(b) Specific requirements.}
The seeker needs a backpack with two separate shoulder straps, a cross-chest strap, and enough space for skates.
LLaVA-v1.6 (\emph{w/} title) recommends a strap-only accessory instead of the full backpack the seeker requested.
LLaVA-v1.6 (\emph{w/} title \& image) does recommend a backpack, but lacks the required strap configuration.
In contrast, \method identifies a backpack with two distinct shoulder straps and a cross-chest strap, leveraging the visual information in the images.

\section{Related Work}
\label{sec:related}

\para{Conversational recommendation.}
Early conversational recommender systems (CRS) rely on handcrafted dialogue flows and critiquing mechanisms.
Users iteratively refine recommendations by providing feedback on item attributes (e.g., \emph{``show me something with a lower price and longer battery life''}).
These methods are largely template or rule-based~\cite{ChenP12,Christakopoulou16,WuLSS19,Lei0MWHKC20,LeiZ0MWCC20,abs-2112-05197,ZhangWSPWXLP22,HeZ0KDM22}, requiring predefined responses for each user critique, and thus limiting flexibility.
With the advent of neural approaches, CRS begins to leverage natural language understanding and generation.
Recent models integrate techniques like knowledge graphs, reinforcement learning, and memory networks to conduct multi-turn dialogues that both converse and recommend items.
Such systems generate more fluent, context-aware responses than rigid templates.
For example, KBRD~\cite{ChenLZDCYT19}, KGSF~\cite{ZhouZBZWY20}, and UniCRS~\cite{WangZWZ22} incorporate external knowledge to enrich the dialogue, and TSCR~\cite{ZouKRRSL22} uses transformer-based architectures for better context understanding.
Recently, large language models (LLMs) have been actively applied to CRS~\cite{abs-2305-07961,WangTZWW23,HeXJSLFMKM23}.
In particular, previous analyses show that LLMs can outperform specialized dialogue recommender systems even without fine-tuning~\cite{HeXJSLFMKM23}.
These analyses show the promise of using LLMs as the backbone of CRS.
However, all of these efforts predominantly handle textual interactions.
Our approach is among the first to incorporate image content directly into CRS, enabling recommendations based on visual preferences.
In contrast to previous CRS that do not consider images, we integrate visual information directly into the model, allowing us to understand and respond to the visual preferences of users.

\para{Visually-aware recommendation.}
Visually-aware recommender systems (VARS) incorporate item images to capture style, design, or aesthetic aspects relevant to user preferences~\cite{LiuHXZGWLT25}.
Early work integrated features extracted by pre-trained convolutional neural networks (CNNs) into collaborative filtering, improving recommendations in domains such as fashion, where visual features play a critical role~\cite{McAuleyTSH15,HeM16a,KangFWM17,LiuWW17,WangWTSRL17}.
Later studies developed end-to-end pipelines that jointly optimized user-item interactions and image representation learning~\cite{HeM16b,WeiWN0HC19}.
In parallel, various architectures have been explored, including graph neural networks~\cite{WeiWN0HC19,WangWYWSN23} and attention-based models~\cite{ChenCXZ0QZ19,LiuCSWNK19,Zhang00WWW21}.
Recently, large vision-language models (VLMs) such as CLIP~\cite{RadfordKHRGASAM21} and VLMo~\cite{BaoW0LMASPW22} have extended these efforts by mapping images and text into a shared embedding space~\cite{abs-2310-20343,abs-2402-08670}.
For example, Rec-GPT4V~\cite{abs-2402-08670} leverages GPT-4Vision~\cite{abs-2309-17421} for zero-shot item ranking by prompting the model with each product’s image. However, this approach can become inefficient for multi-item scenarios due to the high token overhead.
In contrast, our method encodes each product image into a small set of embeddings, retaining essential visual information without incurring excessive token costs.
Our design is well-suited to visually-aware conversational recommendation, where systems must efficiently handle both textual dialogue and multiple product images.

\section{Conclusion and Discussion}
\label{sec:conclusion}

We introduced \method, a two-stage method for visually-aware conversational recommendation.
First, our \emph{visual knowledge self-distillation} compresses each product image into a small set of \texttt{[CLS]}-positioned embeddings while retaining essential visual details.
Second, our \emph{recommendation prompt tuning} enables a large vision-language model to integrate these compressed image embeddings with user dialogues in a unified generative framework.
Experiments on the newly collected \emph{Reddit-Amazon} dataset show substantial performance gains over text-only methods, existing vision-language baselines, and even some proprietary systems.

% Although we examined multi-domain training, merging Beauty, Fashion, and Home offered limited benefit because the domains share little visual or textual knowledge and require domain-specific hyperparameters.
Future work can explore larger or more diverse datasets, where cross-domain attributes may strengthen generalization.
Meanwhile, we considered each product’s \emph{single} representative image (split into sub-images).
Although sufficient in many cases, real-world listings often contain multiple images highlighting different features, suggesting further research on managing richer visual contexts.
Our method follows a candidate-based pipeline, meaning its accuracy partly depends on the retrieval module.
Enhancing retrieval could further boost recommendation performance.
Lastly, this work used a 7B-parameter backbone for computational feasibility.
Larger models (e.g., 13B or 34B) could potentially yield stronger multimodal reasoning at the cost of increased inference time.
Exploring these trade-offs remains an intriguing direction for future efforts.

% \red{
% Discussion (if we have a room):

%     (1) Combined dataset is not very benificial because there are no shared knowledge (i.e., quite different domains) and the required hyperparameters are different? -- maybe need more datasets? -- remain it as our future work.

%     (2) Using multiple images? -- In most cases, the representative image is informative enough, but can remain it as our future work.

%     (3) Performance is essentially dependent on the retrieval.

%     (4) Larger models (13B, 34B, ...) could be more powerful.

% }

\newpage

\bibliographystyle{ACM-Reference-Format}
\balance
\bibliography{paper}

% \clearpage
\appendix
\section{Prompts}
\label{app:prompt}

Our data collection and model training involve several prompt templates, summarized below.
In particular, Table~\ref{tab:filtering_prompt} shows the prompt used to filter Reddit conversations for recommendation requests, while Table~\ref{tab:description_prompt} describes how we generate product-specific image descriptions for visual knowledge self-distillation.
Tables~\ref{tab:context_prompt_wo_image} and~\ref{tab:context_prompt_w_image} illustrate the prompts for the generative recommendation task in both text-only and text \& image settings, respectively.

\para{Filtering recommendation conversations.}
We use GPT-3.5-turbo with a simple instruction (Table~\ref{tab:filtering_prompt}) to identify whether the last utterance in a Reddit thread contains an explicit product recommendation.
Only conversations labeled as `yes' are retained in our \emph{Reddit-Amazon} dataset.

\para{Generating image descriptions.}
For vision self-distillation (Section~\ref{subsec:distillation}), we guide the model to produce a visually grounded description of each product (Table~\ref{tab:description_prompt}).
The prompt encourages the model to focus on appearance and features visible in the product image without simply restating the product title.

\para{Recommendation task with or without images.} Tables~\ref{tab:context_prompt_wo_image} and~\ref{tab:context_prompt_w_image} display the prompt templates for text-only and visually enhanced recommendation.
Both prompts ask the model to read a conversation and a list of candidate products, then recommend the most relevant item by returning its ID.
In the image-based version, each candidate includes an additional image input for multimodal reasoning.

\begin{table}[H]
    \centering
    \caption{
        Prompt for filtering recommendation-related conversations. Our \emph{Reddit-Amazon} dataset includes only conversations that GPT-3.5-turbo~\cite{openai2022chatgpt} answered `yes'.
    }
    \begin{tcolorbox}[
        colback=gray!10,
        colframe=black,
        width=\columnwidth,
        arc=4pt,
        boxrule=1pt,
        ]
    You are a helpful assistant.
    I will show you a conversation on Reddit. Each utterance is included in a tag <utterance>...</utterance> and numbered as (1), (2), (3), etc (i.e., the format is <utterance>(i) [content]</utterance>).
    The goal is to determine whether the final utterance in the conversation is recommending an item.
    If any of the preceding utterances involve requesting an item recommendation, and the last utterance explicitly recommends one or more items, classify the conversation as a recommendation.
    Please answer `yes' if the last utterance is a recommendation of items, otherwise please answer `no'.
    Please answer `uncertain' if you cannot classify `yes' or `no' from the given conversation.
    Do not include other phrases in your answer.

    Conversation:
    
    \blue{\textbf{\{conversation\}}}
    
    \end{tcolorbox}
    \label{tab:filtering_prompt}
\end{table}

\begin{table}[H]
    \centering
    \caption{
        Prompt for generating image description.
    }
    \begin{tcolorbox}[
        colback=gray!10,
        colframe=black,
        width=\columnwidth,
        arc=4pt,
        boxrule=1pt,
        ]    
    You are a helpful AI assistant.
    Given an Amazon product's title and its image, please provide a detailed, visually grounded description of the product that would help someone decide whether to purchase it. Please focus on the product's appearance, features, and any other visually informative aspects. Do not mention the product's title in your answer. 
    
    This product's title is:
    
    \blue{\textbf{\{title\}}}
    
    \blue{\textbf{\{image\}}}
    
    \end{tcolorbox}
    \label{tab:description_prompt}
\end{table}

\begin{table}[H]
    \centering
    \caption{
        Prompt for recommendation used by generative baselines (\emph{w/o} images) and \method.
    }
    \begin{tcolorbox}[
        colback=gray!10,
        colframe=black,
        width=\columnwidth,
        arc=4pt,
        boxrule=1pt,
        ]

    You are an AI assistant specialized in providing personalized product recommendations based on user conversations. You are given a conversation between a user seeking recommendation (denoted by $\langle$submission$\rangle$) and other users providing comments (denoted by $\langle$comment$\rangle$). You are also given a set of candidate products with their IDs and titles formatting as ``ID: title''. Among the candidates, recommend the most relevant product to the seeker. Only reply with its ID, and don't say anything else.

    Conversation:
    
    \blue{\textbf{\{conversation\}}}

    Candidates:

    \blue{\textbf{\{list of candidates\}}}
    
    \end{tcolorbox}
    \label{tab:context_prompt_wo_image}
\end{table}

\begin{table}[H]
    \centering
    \caption{
        Prompt for recommendation used by generative baselines (\emph{w/} images) and \method.
    }
    \begin{tcolorbox}[
        colback=gray!10,
        colframe=black,
        width=\columnwidth,
        arc=4pt,
        boxrule=1pt,
        ]

    You are an AI assistant specialized in providing personalized product recommendations based on user conversations. You are given a conversation between a user seeking recommendation (denoted by $\langle$submission$\rangle$) and other users providing comments (denoted by $\langle$comment$\rangle$). You are also given a set of candidate products with their IDs, titles, and images formatting as ``ID: title'' followed by an image. Among the candidates, recommend the most relevant product to the seeker. Only reply with its ID, and don't say anything else.

    Conversation:
    
    \blue{\textbf{\{conversation\}}}

    Candidates:

    \blue{\textbf{\{list of candidates\}}}
    
    \end{tcolorbox}
    \label{tab:context_prompt_w_image}
\end{table}

\section{Hyperparameters and Training Details}
\label{app:hyperparams}

\para{Visual knowledge self-distillation (Section~\ref{subsec:distillation}).}
During the generation process, we set the maximum output length to 128 tokens.
For the distillation process, we use a batch size of $4$ and apply LoRA~\cite{HuSWALWWC22} to the vision tower and projector with $r=8$, $\alpha=32$, and a $0.1$ dropout rate.
We explore learning rates in $\{5\times10^{-5},\,10^{-5},\,5\times10^{-6},\,10^{-6}\}$ and weight decays in $\{10^{-2},\,10^{-3},\,10^{-4},\,10^{-5},\,0\}$, training for up to $5$ epochs and picking the best checkpoint based on validation performance.
In practice, results typically peak around epoch $2$.

\para{Recommendation prompt tuning (Section~\ref{subsec:prompt_tuning}).}
We limit the input length to 2K tokens, set the batch size to $1$, and again apply LoRA with $r=8$, $\alpha=32$, and a $0.1$ dropout rate.
We repeat the same grid search for learning rates and weight decays, training for up to $5$ epochs and selecting the best model by validation performance.
Most models converge by epoch $1$ or $2$ under these conditions.

\section{Additional Experiments}
\label{app:experiments}
Tables~\ref{tab:comparison_open_openai} and~\ref{tab:comparison_close_openai} report experimental results when using OpenAI-emb\textsubscript{large} as the retrieval instead of SBERT.
We compare our method (\method) against both open-source and proprietary baselines in text-only (item title) and combining text and image (item title and image) settings.

Table~\ref{tab:comparison_open_openai} compares \method with open-source models.
OpenAI-emb\textsubscript{large} greatly improves initial retrieval quality when it is compared to SBERT (Table~\ref{tab:comparison_open_sbert}).
In particular, even the generative baselines underperform this retrieval model.
Nevertheless, \method maintains a performance advantage across all domains (beauty, fashion, and home), outperforming the strongest open-source competitor by up to 28.3\% in the fashion domain.
When coupled with images, other methods often struggle to effectively process multiple images and exhibit lower accuracy.

Table~\ref{tab:comparison_close_openai} compares \method with GPT-3.5-turbo, GPT-4o-mini, and GPT-4o under using OpenAI-emb\textsubscript{large} as the retrieval.
GPT-4o still achieves the highest overall HR@1.
However, \method remains competitive in all domains and surpasses GPT-3.5-turbo.
These results show that the design of \method to address token explosion preserves strong recommendation accuracy, even when a more powerful retrieval approach is used.

\begin{table}[H]
    \setlength\tabcolsep{3.5pt}
    \small
    \centering
    \caption{
         Performance comparison of \method with open-source baselines.
        \colorbox{gray!10}{Gray colored} methods require a candidate retrieval to generate answers.
        We use OpenAI-emb\textsubscript{large}, which is one of the most powerful proprietary encoder, as the common retrieval.
        Bold and underline indicate the best and the second-best, respectively.
    }
    \begin{tabular}{lcccccc}
        \toprule
        \multirow{2}{*}{\textbf{Method}} & \multicolumn{2}{c}{\textit{\textbf{Beauty}}} & \multicolumn{2}{c}{\textit{\textbf{Fashion}}} & \multicolumn{2}{c}{\textit{\textbf{Home}}}\\
        \cmidrule(lr){2-3} \cmidrule(lr){4-5} \cmidrule(lr){6-7}
        & \textbf{HR@1} & \textbf{VR} & \textbf{HR@1} & \textbf{VR} & \textbf{HR@1} & \textbf{VR}\\
        \midrule
        \multicolumn{7}{c}{\textit{Retrieval Baselines (item title)}}\\
        BM25 & 0.0169 & - & 0.0140 & - & 0.0479 & -\\
        SBERT & 0.0551 & - & 0.0681 & - & 0.2166 & -\\
        RoBERTa\textsubscript{large} & 0.0640 & - & 0.0631 & - & 0.1814 & -\\
        SimCSE\textsubscript{large} & 0.0326 & - & 0.0301 & - & 0.0957 & -\\
        BLaIR\textsubscript{base} & 0.0371 & - & 0.0441 & - & 0.1335 & -\\
        OpenAI-emb\textsubscript{large} & \underline{0.1461} & - & \underline{0.1393} & - & \underline{0.3224} & -\\
        \midrule
        \rowcolor{gray!10}\multicolumn{7}{c}{\textit{Generative Baselines (item title)} + OpenAI-emb\textsubscript{large}}\\
        \rowcolor{gray!10}Vicuna-v1.5 & 0.0728 & 0.9819 & 0.0718 & 0.9867 & 0.1360 & 1.0000 \\
        \rowcolor{gray!10}LLaVA-v1.5 & 0.0762 & 0.9832 & 0.0768 & 0.9903 & 0.1114 & 0.9946 \\
        \rowcolor{gray!10}LLaVA-v1.6 & 0.0972 & 0.9870 & 0.1150 & 0.9867 & 0.2430 & 0.9839 \\
        \midrule
        \rowcolor{gray!10}\multicolumn{7}{c}{\textit{Generative Baselines (item title and image)} + OpenAI-emb\textsubscript{large}}\\
        \rowcolor{gray!10}LLaVA-v1.5 & 0.0000 & 0.0000 & 0.0000 & 0.0000 & 0.0000 & 0.0000\\
        \rowcolor{gray!10}LLaVA-v1.6 & 0.0641 & 0.9728 & 0.0857 & 0.9806 & 0.1646 & 0.9946\\
        \midrule
        \rowcolor{gray!10}\multicolumn{7}{c}{\textit{Proposed Method} + OpenAI-emb\textsubscript{large}}\\
        \rowcolor{gray!10}\textbf{\method (ours)} & \textbf{0.1743} & 0.9676 & \textbf{0.1787} & 0.9455 & \textbf{0.3537} & 0.9892\\
        \rowcolor{cyan!20}\textbf{Improvement} & \textbf{+19.3\%} & - & \textbf{+28.3\%} & - & \textbf{+9.7\%} & -\\
        \bottomrule
    \end{tabular}
    \label{tab:comparison_open_openai}
\end{table}

\begin{table}[H]
    \setlength\tabcolsep{4.2pt}
    \small
    \centering
    \caption{
        Performance comparison of \method with proprietary baselines.
        We use OpenAI-emb\textsubscript{large} as a candidate retrieval.
    }
    \begin{tabular}{lcccccc}
        \toprule
        \multirow{2}{*}{\textbf{Method}} & \multicolumn{2}{c}{\textit{\textbf{Beauty}}} & \multicolumn{2}{c}{\textit{\textbf{Fashion}}} & \multicolumn{2}{c}{\textit{\textbf{Home}}}\\
        \cmidrule(lr){2-3} \cmidrule(lr){4-5} \cmidrule(lr){6-7}
        & \textbf{HR@1} & \textbf{VR} & \textbf{HR@1} & \textbf{VR} & \textbf{HR@1} & \textbf{VR}\\
        \midrule
        \multicolumn{7}{c}{\textit{Generative Baselines (item title)} + OpenAI-emb\textsubscript{large}}\\
        GPT-3.5-turbo & 0.1449 & 0.9987 & 0.1523 & 0.9964 & 0.2997 & 1.0000 \\
        GPT-4o-mini & 0.1809 & 1.0000 & 0.1755 & 0.9976 & 0.3552 & 1.0000\\
        GPT-4o & 0.2005 & 0.9948 & 0.1944 & 0.9939 & 0.4055 & 1.0000 \\
        \midrule
        \multicolumn{7}{c}{\textit{Generative Baselines (item title and image)} + OpenAI-emb\textsubscript{large}}\\
        GPT-4o-mini & 0.1667 & 0.9974 & 0.1623 & 0.9976 & 0.3526 & 1.0000\\
        GPT-4o & 0.1914 & 0.9974 & 0.1942 & 0.9939 & 0.3980 & 1.0000\\
        \midrule
        \multicolumn{7}{c}{\textit{Proposed Method} + OpenAI-emb\textsubscript{large}}\\
        \textbf{\method (ours)} & 0.1743 & 0.9676 & 0.1787 & 0.9455 & 0.3537 & 0.9892\\
        \bottomrule
    \end{tabular}
    \label{tab:comparison_close_openai}
\end{table}

\section{Datasets for Conversational Recommendation}
\label{app:dataset_convrec}

Table~\ref{tab:convrec_datasets} compares well-known conversational recommendation datasets with our newly introduced \textit{Reddit-Amazon} dataset.
Early efforts (e.g., ReDial~\cite{LiKSMCP18}, GoRecDial~\cite{KangBSCBW19}, and TG-ReDial~\cite{ZhouZZWW20}) primarily focus on movies, often with crowd-sourced or synthetic dialogues.
Recent datasets have expanded into broader domains (e.g., music, restaurants, and e-commerce) or used natural Reddit conversations~\cite{HeXJSLFMKM23,00010DFWFCZCLC23,JiaLW0XLSCP022}, but still lack substantial visual information.
In contrast, our \textit{Reddit-Amazon} covers three visually oriented categories (beauty, fashion, and home), collecting 19K naturally occurring conversations with 51K turns.
It aligns each conversation with 15K unique items and their images, capturing realistic discussions where users frequently refer to product appearances.
This emphasis on visual details fills an important gap, allowing more comprehensive evaluations of visually-aware conversational recommender systems.

\begin{table*}[t]
    % \setlength\tabcolsep{0.5pt}
    % \small
    \centering
    \caption{
        Comparison of existing conversational recommendation datasets. Our \emph{Reddit-Amazon} dataset consists of realistic conversations and focuses on visual-oriented domains such as beauty, fashion, and home.
    }
    \begin{tabular}{lrrrrl}
        \toprule
        \textbf{Datasets} & \textbf{\#Conv.} & \textbf{\#Turns} & \textbf{\#Items} & \textbf{Domain} & \textbf{Source} \\
        \midrule
        FacebookRec~\cite{DodgeGZBCMSW15} & 1M & 6M & - & Movies & Synthetic\\
        ReDial~\cite{LiKSMCP18} & 10K & 182K & 6.2K & Movies & Crowd-sourced\\
        GoRecDial~\cite{KangBSCBW19} & 9K & 170K & - & Movies & Crowd-sourced\\
        OpenDialKG~\cite{MoonSKS19} & 15K & 91K & - & Movies, music, etc. & Crowd-sourced\\
        TG-ReDial~\cite{ZhouZZWW20} & 10K & 129K & - & Movies & Synthetic\\
        DuRecDial 2.0~\cite{Liu0N0C21} & 16.5K & 255K & - & Movies, music, etc. & Crowd-sourced\\
        CCPE-M~\cite{RadlinskiBBK19} & 502 & 11K & - & Movies & Crowd-sourced\\
        INSPIRED~\cite{HayatiKZSY20} & 1K & 35K & 1.9K & Movies & Crowd-sourced\\
        Reddit-Movie\textsubscript{base}~\cite{HeXJSLFMKM23} & 85K & 133K & 24.3K & Movies & Natural\\
        Reddit-Movie\textsubscript{large}~\cite{HeXJSLFMKM23} & 634K & 1.6M & 51.2K & Movies & Natural\\
        U-NEED~\cite{00010DFWFCZCLC23} & 7K & 53K & - & E-commerce & Natural\\
        E-ConvRec~\cite{JiaLW0XLSCP022} & 25K & 775K & - & E-commerce & Natural\\
        HOOPS~\cite{FuXZXLMZ21} & - & 11.6M & - & E-commerce & Synthetic\\
        MGConvRec~\cite{XuMLLSY20} & 7K & 73K & - & Restaurant & Crowd-sourced\\
        MMConv~\cite{LiaoLZHC21} & 5K & 39K & - & Travel & Crowd-sourced\\
        MobileConvRec~\cite{abs-2405-17740} & 12.2K & 156K & 1.7K & Music, sports, etc. & Synthetic\\
        \midrule
        \textbf{Reddit-Amazon (ours)} & 19K & 51K & 15K & Beauty, fashion, home & Natural\\
        \bottomrule
    \end{tabular}
    \label{tab:convrec_datasets}
\end{table*}

\section{Examples of Image Descriptions from LLaVA-v1.6}
\label{app:generation_examples}

Tables~\ref{tab:generated_beauty},~\ref{tab:generated_fashion}, and~\ref{tab:generated_home} show examples of image descriptions generated by LLaVA-v1.6 in our visual knowledge self-distillation step.
These examples are drawn from three domains in the \emph{Reddit-Amazon} dataset: \emph{beauty}, \emph{fashion}, and \emph{home}, each table corresponding to one domain.
For each product, we provide its image, domain label, product title, and the description automatically generated by LLaVA-v1.6.

We highlight in yellow shading specific visual details inferred from the image, such as color, material, design elements, or functional components.
These features are not explicitly mentioned in the product title, demonstrating the model’s capacity to generate visually grounded information.
In our \emph{visual knowledge self-distillation} process (Section~\ref{subsec:distillation}), we distill only a small set of tokens to capture this descriptive knowledge for subsequent recommendation tasks.
This approach preserves the essential details found in the images while mitigating token overflows when multiple candidates must be processed simultaneously.

\begin{table*}[t]
    \centering
    \caption{
        Examples of image descriptions generated by LLaVA-v1.6 in the \emph{beauty} domain. Highlighted text indicates additional visual details inferred from the image that are not explicitly mentioned in the product title.
    }
    \begin{tabular}{m{0.2\textwidth} m{0.1\textwidth} m{0.2\textwidth} m{0.4\textwidth}}
        \toprule
        \textbf{Image} & \textbf{Domain} & \textbf{Title} & \textbf{Description generated by LLaVA-v1.6}\\
        \midrule
        \includegraphics[width=0.18\textwidth]{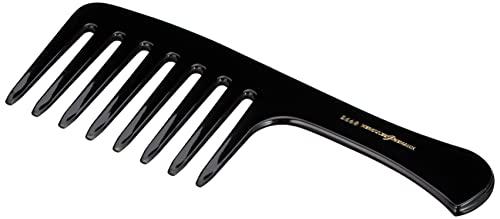} & Beauty & Hercules Sägemann, Magic Star comb, 9 inches, 1 set, (1 x 1 piece), 5660 &
        The product in the image is a wide-tooth comb, specifically the Hercules Sägemann Magic Star comb. It is 9 inches in length and \hl{appears to be made of a durable material, likely plastic or a similar synthetic}. The comb has \hl{a sleek, black design with wide, evenly spaced teeth}, making it suitable for \hl{detangling hair without causing damage}. The handle is \hl{ergonomically designed for a comfortable grip}, and the comb features the brand name ``Hercules Sägemann'' printed on it. \hl{This type of comb is ideal for use on wet or dry hair and is particularly useful for individuals with curly or thick hair.}\\
        \midrule
        \includegraphics[width=0.18\textwidth]{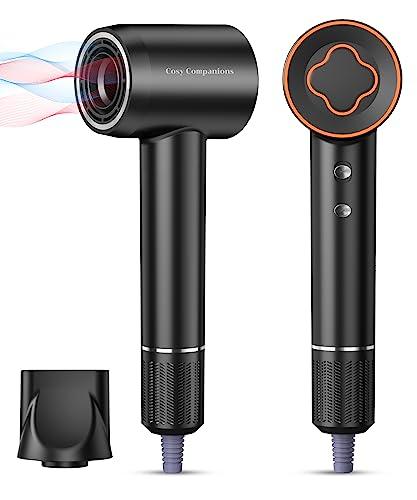} & Beauty & Hair Dryer - 150000 RPM High-Speed Brushless Motor Negative Ionic Blow Dryer for Fast Drying, Low Noise Thermo-Control Hair Dryers with Magnetic Nozzle, Perfect for Gifts &
        Based on the image, the product is a sleek, modern hair dryer with a high-speed brushless motor. The hair dryer has \hl{a cylindrical design with a matte black finish and silver accents}. It features \hl{a magnetic nozzle attachment for focused airflow}, which can be seen detached in the image. The front of the hair dryer shows the air outlet, which appears to be designed for efficient airflow, with blue and red lines indicating the flow of air.
        
        The handle of the hair dryer has \hl{a textured grip for easy handling and two buttons, likely for controlling the speed and heat settings}. The back of the hair dryer has a unique design with \hl{an orange ring and a cross-shaped pattern}, possibly indicating the air intake area.
        
        Overall, the hair dryer looks compact and stylish, designed for fast drying with low noise, and it includes advanced features like negative ionic technology and thermo-control. The product appears to be suitable for gifting, as suggested by its elegant design and high-tech features.\\
        \midrule
        \includegraphics[width=0.18\textwidth]{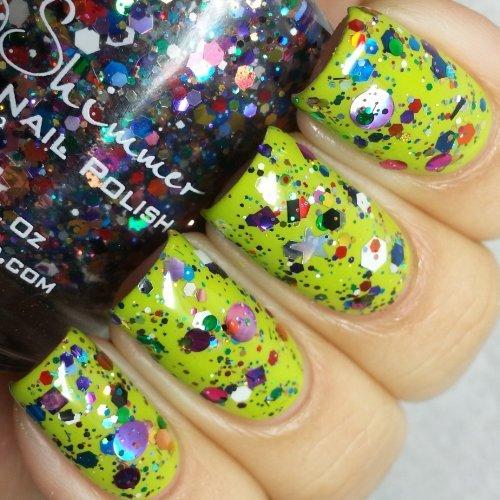} & Beauty & Clown Puke Nail Polish - 0.5 Oz Full Sized Bottle &
        The ``Clown Puke Nail Polish'' \hl{appears to be a vibrant and playful nail polish}. The image shows nails painted with \hl{a bright, lime green base color, which is then covered with a variety of colorful, multi-shaped glitter pieces}. The glitter includes \hl{hexagons, stars, and circles in various colors such as blue, red, purple, and white}. The overall effect is a fun, confetti-like appearance that is eye-catching and festive. The nail polish comes in a 0.5 oz full-sized bottle.\\
        \bottomrule
    \end{tabular}
    \label{tab:generated_beauty}
\end{table*}

\begin{table*}[t]
    \centering
    \caption{
        Examples of image descriptions generated by LLaVA-v1.6 in the \emph{fashion} domain. Highlighted text indicates additional visual details inferred from the image that are not explicitly mentioned in the product title.
    }
    \begin{tabular}{m{0.2\textwidth} m{0.1\textwidth} m{0.2\textwidth} m{0.4\textwidth}}
        \toprule
        \textbf{Image} & \textbf{Domain} & \textbf{Title} & \textbf{Description generated by LLaVA-v1.6}\\
        \midrule
        \includegraphics[width=0.18\textwidth]{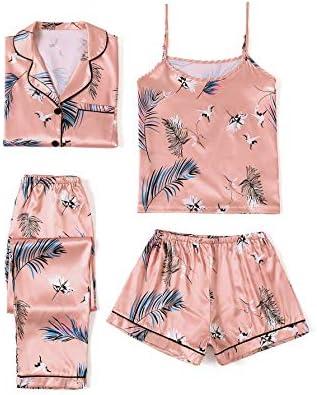} & Fashion & LYANER Women's Pajamas Set 4pcs Satin Silk Cami Top Button Down Loungewear Pjs Set &
        The product is a women's pajama set made of satin silk, consisting of four pieces. The set includes: 
        
        1. A button-down \hl{long-sleeve top with a collar and black piping details}.
        
        2. A cami top \hl{with thin spaghetti straps}.
        
        3. \hl{A pair of long pants with an elastic waistband.}
        
        4. \hl{A pair of shorts with an elastic waistband.}
        
        The fabric has \hl{a smooth, shiny finish} and features \hl{a pink base color with a tropical print, including leaves and flowers in shades of blue, white, and black}. The overall design is \hl{elegant and comfortable, suitable for lounging and sleeping}.\\
        \midrule
        \includegraphics[width=0.18\textwidth]{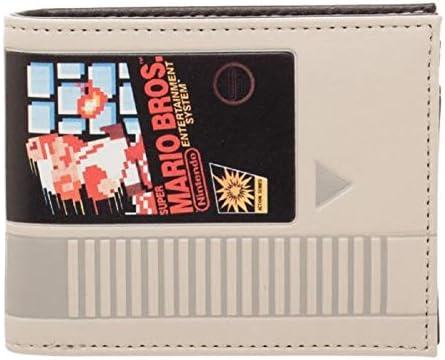} & Fashion & Super Mario Bros Classic Retro NES Design Bi-fold Wallet &
        The product is a bi-fold wallet designed with a retro theme inspired by the classic Super Mario Bros game for the NES (Nintendo Entertainment System). The wallet features \hl{a design that mimics the appearance of an NES game cartridge}, with the iconic Super Mario Bros label prominently displayed on the front. The label includes \hl{pixelated graphics reminiscent of the original game, along with the Nintendo logo and other classic design elements}. The wallet appears to be made of \hl{a material that resembles the texture and color of an NES cartridge, giving it a nostalgic and vintage look}.\\
        \midrule
        \includegraphics[width=0.18\textwidth]{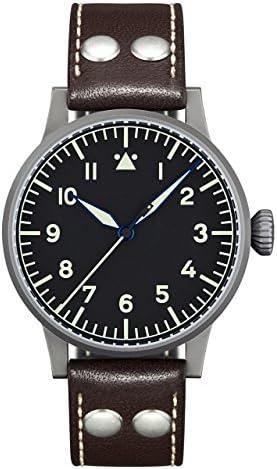} & Fashion & Laco Munster Type A Dial Swiss Automatic Pilot Watch with Sapphire Crystal 861748 &
        The product in the image is a wristwatch. Here are some key features based on the image:
        
        1. **Dial**: \hl{The watch has a black dial with white Arabic numerals and markers. The numerals are in a classic pilot watch style, with large, easy-to-read numbers.}
        
        2. **Hands**: \hl{The watch features blue-colored hands, with the hour and minute hands being sword-shaped and the second hand being a thin, straight line.}
        
        3. **Case**: \hl{The case appears to be made of stainless steel with a matte finish.}
        
        4. **Crystal**: The watch has \hl{a clear, flat} sapphire crystal covering the dial.
        
        5. **Crown**: \hl{The crown is located at the 3 o'clock position and has a textured grip for easy adjustment.}
        
        6. **Strap**: \hl{The watch has a brown leather strap with white stitching and two large rivets near the lugs, which is characteristic of traditional pilot watches.}
        
        7. **Design**: \hl{The overall design is reminiscent of classic pilot watches, with a focus on legibility and functionality.}
        
        This watch combines a vintage pilot watch aesthetic with modern materials and craftsmanship.\\
        \bottomrule
    \end{tabular}
    \label{tab:generated_fashion}
\end{table*}

\begin{table*}[t]
    \centering
    \caption{
        Examples of image descriptions generated by LLaVA-v1.6 in the \emph{home} domain. Highlighted text indicates additional visual details inferred from the image that are not explicitly mentioned in the product title.
    }
    \begin{tabular}{m{0.2\textwidth} m{0.1\textwidth} m{0.2\textwidth} m{0.4\textwidth}}
        \toprule
        \textbf{Image} & \textbf{Domain} & \textbf{Title} & \textbf{Description generated by LLaVA-v1.6}\\
        \midrule
        \includegraphics[width=0.18\textwidth]{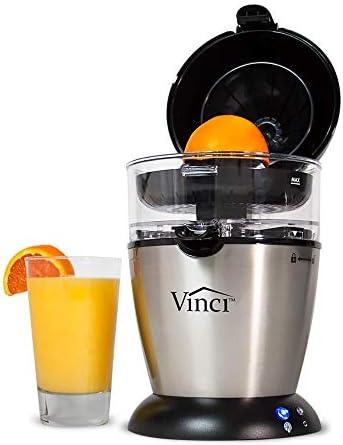} & Home & Vinci Hands-Free Patented Electric Citrus Juicer 1-Button Easy Press Lemon Lime Orange Grapefruit Juice Squeezer Easy to Clean Juicer Machine, Black/Stainless Steel &
        The product in the image is an electric citrus juicer by Vinci. It features a sleek design with a combination of black and stainless steel colors. The juicer has \hl{a transparent upper section} where the citrus fruit is placed, and it appears to be in the process of juicing an orange. \hl{The machine has a one-button operation for easy use, as indicated by the buttons on the base. The base is black and has a sturdy appearance.} Next to the juicer, there is a glass filled with freshly squeezed orange juice, garnished with a slice of orange, showcasing the juicer's capability to produce fresh juice. The overall design suggests that the juicer is easy to clean and user-friendly.\\
        \midrule
        \includegraphics[width=0.18\textwidth]{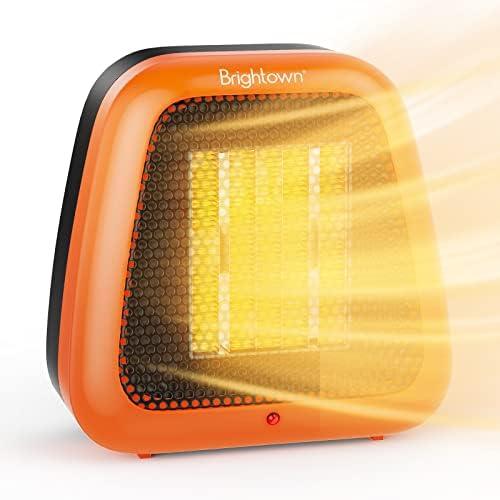} & Home & Small Space Heater for Indoor Use - 400W Low Wattage Mini Heater with Overheat Protection, Portable Personal Heater with Tip Over Protection, Low Noise Desk Heater for Office Bedroom Home Use &
        Based on the image, the product is a compact, portable space heater designed for indoor use. \hl{It has a modern, triangular design with an orange front and a black back. The front features a mesh grille through which the heating elements are visible, and it appears to be emitting warm air.} The heater is branded with the name \"Brightown\" at the top. It is likely designed for personal use in small spaces such as an office, bedroom, or home. The heater includes safety features such as overheat protection and tip-over protection, and it operates with low noise, making it suitable for quiet environments.\\
        \midrule
        \includegraphics[width=0.18\textwidth]{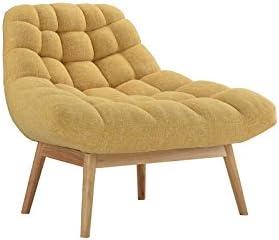} & Home & Modern Linen Plush Living Room Lounge Accent Chair (Yellow) &
        The product in the image is a modern accent chair designed for a living room or lounge area. It features a plush, cushioned seat and backrest upholstered in yellow linen fabric. The chair has \hl{a tufted design}, giving it a stylish and comfortable appearance. \hl{The legs are made of wood, providing a sturdy and natural look that complements the overall modern aesthetic of the chair.}\\
        \bottomrule
    \end{tabular}
    \label{tab:generated_home}
\end{table*}

\end{document}